\documentclass[10pt, conference, compsocconf]{IEEEtran}
\ifCLASSINFOpdf
\else
\fi
\usepackage[bookmarks=false]{hyperref}
\usepackage{graphicx}
\usepackage{subfigure}
\usepackage{bbding}
\usepackage{multirow}
\usepackage{amsmath}
\usepackage{amssymb}
\usepackage{epstopdf}
\usepackage{url}
\usepackage{amsthm}

\usepackage{booktabs}
\usepackage{algorithm}
\usepackage{algorithmic}
\usepackage{diagbox}
\usepackage{chngpage}
\usepackage{graphicx}
\usepackage{graphics}
\usepackage[normalem]{ulem}
\usepackage{soul}
\usepackage{cancel}

\newcommand*{\Scale}[2][4]{\scalebox{#1}{$#2$}}%

\begin{document}
%
\title{Balanced Distribution Adaptation for Transfer Learning}


\author{\IEEEauthorblockN{
		Jindong Wang\IEEEauthorrefmark{1}\IEEEauthorrefmark{2}\IEEEauthorrefmark{3},
		Yiqiang Chen\IEEEauthorrefmark{1}\IEEEauthorrefmark{2}\IEEEauthorrefmark{3},
		Shuji Hao\IEEEauthorrefmark{4}, 
		Wenjie Feng\IEEEauthorrefmark{1}\IEEEauthorrefmark{3}\IEEEauthorrefmark{8},
		Zhiqi Shen\IEEEauthorrefmark{6}
	}
	\IEEEauthorblockA{\IEEEauthorrefmark{1}Beijing Key Laboratory of Mobile Computing and Pervasive Device}
	\IEEEauthorblockA{\IEEEauthorrefmark{8}CAS Key Laboratory of Network Data Science \& Technology}
	\IEEEauthorblockA{\IEEEauthorrefmark{2}Institute of Computing Technology, Chinese Academy of Sciences, Beijing, China}
	\IEEEauthorblockA{\IEEEauthorrefmark{3}University of Chinese Academy of Sciences,\IEEEauthorrefmark{4}Institute of High Performance Computing, A*STAR}
	\IEEEauthorblockA{\IEEEauthorrefmark{6}School of Computer Science and Engineering, Nanyang Technological University, Singapore}
	Email:\{wangjindong,yqchen\}@ict.ac.cn, haosj@ihpc.a-star.edu.sg, fengwenjie@software.ict.ac.cn, zqshen@ntu.edu.sg}


%


\maketitle

\begin{abstract}
Transfer learning has achieved promising results by leveraging knowledge from the source domain to annotate the target domain which has few or none labels. Existing methods often seek to minimize the distribution divergence between domains, such as the marginal distribution, the conditional distribution or both. However, these two distances are often treated equally in existing algorithms, which will result in poor performance in real applications. Moreover, existing methods usually assume that the dataset is balanced, which also limits their performances on imbalanced tasks that are quite common in real problems. To tackle the distribution adaptation problem, in this paper, we propose a novel transfer learning approach, named as \underline{B}alanced \underline{D}istribution \underline{A}daptation~(BDA), which can adaptively leverage the importance of the marginal and conditional distribution discrepancies, and several existing methods can be treated as special cases of BDA. Based on BDA, we also propose a novel \underline{W}eighted \underline{B}alanced \underline{D}istribution \underline{A}daptation~(W-BDA) algorithm to tackle the class imbalance issue in transfer learning. W-BDA not only considers the distribution adaptation between domains but also adaptively changes the weight of each class. To evaluate the proposed methods, we conduct extensive experiments on several transfer learning tasks, which demonstrate the effectiveness of our proposed algorithms over several state-of-the-art methods.

\end{abstract}


%
\IEEEpeerreviewmaketitle

\section{Introduction}
\label{sec-intro}

Preparing labeled data is crucial for training machine learning algorithms. However, it is often expensive and time-consuming to obtain sufficient labeled data in real applications. In this case, transfer learning~\cite{pan2010survey} has been a promising approach by transferring knowledge from a labeled source domain to the target domain. Transfer learning often assumes the training and testing data are from similar but different distributions~\cite{pan2010survey}. For instance, the images of an object taken in different angles, backgrounds and illuminations could lead to different marginal or conditional distributions. By observing this, existing transfer learning methods are mainly focusing on \textit{distribution adaptation} to minimize the distribution divergence between domains~\cite{long2013transfer,yan2017mind,hou2016unsupervised}.



Most of the existing distribution adaptation methods adapt either marginal distribution~\cite{pan2011domain}, conditional distribution~\cite{satpal2007domain} or both~\cite{long2013transfer,hou2016unsupervised}. It is shown in~\cite{long2013transfer} that adapting both could achieve better performance. The work of~\cite{long2013transfer,hou2016unsupervised,tahmoresnezhad2016visual} also proposed several approaches to adapt the joint distribution. However, those two distributions are often treated equally in existing methods, while the importance of each other is not leveraged. When the datasets are much more dissimilar, it means the marginal distributions are more dominant; when the datasets are similar, it means the conditional distributions needs more attention. Hence, it will deteriorate the performance of algorithms by only adding them together with equal weight. Therefore, how to adaptively leverage the importance of each distribution is a critical problem.

Moreover, class imbalance often exists in many transfer learning scenarios.
When the class proportion of domains is highly imbalanced, it needs to be considered carefully for distribution adaptation. Existing methods~\cite{long2013transfer,hou2016unsupervised} often ignore this issue by treating the classes as balanced across domains, or they only handle the bias on one domain~\cite{yan2017mind}, and this may hinder the effectiveness of transfer learning. Therefore, how to handle the class imbalance situation in transfer learning is another important challenge.

In this paper, we propose two novel methods to tackle the above two issues. For distribution adaptation, we propose \underline{B}alanced \underline{D}istribution \underline{A}daptation~(BDA). BDA can not only adapt both the marginal and conditional distributions between domains, but also leverage the importance of those two distributions, thus it can be effectively adjusted to specific transfer learning tasks. Several existing methods can be regarded as special cases of BDA. Based on BDA, we also propose a novel \underline{W}eighted \underline{B}alanced \underline{D}istribution \underline{A}daptation~(W-BDA) algorithm to tackle the class imbalance issue in transfer learning. The proposed W-BDA can adaptively change the weight of each class when performing distribution adaptation. To evaluate BDA and W-BDA, we conduct extensive experiments on five image datasets.

To sum up, our contributions are mainly three-fold:

1) We propose a novel transfer learning method, which is named as BDA to balance the marginal and conditional distribution adaptation. BDA can adaptively adjust the importance of those two distances and can achieve a better performance. Several transfer learning algorithms can be regarded as special cases of BDA.

2) We also propose another novel method W-BDA by extending BDA to handle the class imbalance problem which is common in transfer learning. The proposed W-BDA not only considers the distribution adaptation of domains but also adaptively changes the weight of each class, thus it can handle the class imbalance problem for transfer learning.

3) We conduct extensive experiments on five image datasets to evaluate the BDA and W-BDA methods, indicating their superiority against other state-of-the-art methods.



\section{Related Work}
\label{sec-related}
Transfer learning has been widely applied to activity recognition~\cite{wang2017deep}, incremental learning~\cite{hu2016less}, and online learning~\cite{hao2017online,chen2016ocean}. Our proposed BDA and W-BDA are mainly related to the feature-based transfer learning methods. Thus, in this section, we present a detailed discussion on this category, specifically on two aspects.

%
%

\textit{Joint distribution adaptation.} \cite{li2016joint} proposed to jointly select feature and preserve structural properties. Long \textit{et al.}~\cite{long2013transfer} proposed joint distribution adaptation method (JDA) to match both marginal and conditional distribution between domains. Others extended JDA by adding structural consistency~\cite{hou2016unsupervised}, domain invariant clustering~\cite{tahmoresnezhad2016visual}, and target selection~\cite{hou2015unsupervised}. Those methods tend to ignore the importance between two distinct distributions by just adding them together. However, when there is a large discrepancy between both distributions, those methods cannot evaluate the importance of each distribution, and may not generalize well in most cases. Our work is capable of investigating the importance of each distribution. Thus it can be more generalized to transfer learning scenarios with complex data distributions.

\textit{Class imbalance problem.} Previous sample re-weighting methods~\cite{ando2017deep} only learned weights of specific samples, but ignore the class weights balance for different classes. \cite{ming2015unsupervised} developed a Closest Common Space Learning (CCSL) method to adapt the cross-domain weights. CCSL is an instance selection method, while ours is a feature based approach. Multiset feature learning was proposed in~\cite{wu2017multiset} to learn discriminant features. \cite{yan2017mind} proposed weighted maximum mean discrepancy to construct a source reference collection on the target domain but it only adapted the prior of source domain, while our method could adapt the priors from both source and target domains. \cite{hsiao2016learning} tackled the imbalance issue when target domain has some labels, while in our method, target domain has no labels. \cite{li2016prediction} adjusted the weights of different samples according to their predictions, while our work focuses on adjusting the weight of each class. 

\section{Balanced Distribution Adaptation}
\label{sec-method}
This section elaborates our proposed algorithms. First, we introduce the problem definition. Then, we present the Balanced Distribution Adaptation~(BDA) approach. Finally, the Weighted BDA~(W-BDA) method is introduced.

\subsection{Problem Definition}
\label{sec-method-problem}

%

Given a labeled source domain $\{\mathbf{x}_{s_i},y_{s_i}\}^n_{i=1}$, an unlabeled target domain $\{\mathbf{x}_{t_j}\}^m_{j=1}$, and assume feature space $\mathcal{X}_s = \mathcal{X}_t$, label space $\mathcal{Y}_s = \mathcal{Y}_t$ but marginal distributions $P_s(\mathbf{x}_s) \ne P_t(\mathbf{x}_t)$ with conditional distributions $P_s(y_s|\mathbf{x}_s) \ne P_s(y_t|\mathbf{x}_t)$. Transfer learning aims to learn the labels $\mathbf{y}_t$ of $\mathcal{D}_{t}$ using the source domain $\mathcal{D}_s$.

Balanced distribution adaptation solves the transfer learning problem by \textit{adaptively} minimizing the marginal and conditional distribution discrepancy between domains, and handle the class imbalance problem, i.e. to minimize the discrepancies between: 1) $P(\mathbf{x}_s)$ and $P(\mathbf{x}_t)$, 2) $P(y_s|\mathbf{x}_s)$ and $P(y_t|\mathbf{x}_t)$.

\subsection{Balanced Distribution Adaptation}
\label{sec-method-bda}

Transfer learning methods often seek to adapt both the marginal and conditional distributions between domains~\cite{long2013transfer,tahmoresnezhad2016visual}. Specifically, this refers to minimizing the distance
\begin{equation}
\label{equ-mmd}
\begin{split}
D(\mathcal{D}_s,\mathcal{D}_t) \approx &\ D(P(\mathbf{x}_s),P(\mathbf{x}_t)) \\
&+ D(P(y_s|\mathbf{x}_s),P(y_t|\mathbf{x}_t))
\end{split}
\end{equation}

However, simply matching both distributions is \textbf{not} enough. Existing methods usually assume they are equally important, and that implicit assumption does not hold. In this section, we propose \underline{B}alanced \underline{D}istribution \underline{A}daptation~(BDA) to adaptively adjust the importance of both the marginal and conditional distributions based on each specific tasks. Concretely speaking, BDA exploits a \textit{balance factor} $\mu$ to leverage the different importance of distributions:
\begin{equation}
\label{equ-mummd}
	\begin{split}
	D(\mathcal{D}_s,\mathcal{D}_t) \approx (1 &- \mu)D(P(\mathbf{x}_s),P(\mathbf{x}_t))\\
	&+ \mu D(P(y_s|\mathbf{x}_s),P(y_t|\mathbf{x}_t))\\
	\end{split}
\end{equation}
where $\mu \in [0,1]$. When $\mu \rightarrow 0$, it means the datasets are more dissimilar, so the marginal distribution is more dominant; when $\mu \rightarrow 1$, it reveals the datasets are similar, so the conditional distribution is more important to adapt. Therefore, the balance factor $\mu$ can adaptively leverage the importance of each distribution and lead to good results.

It is worth noting that, since the target domain $\mathcal{D}_t$ has no labels, it is not feasible to evaluate the conditional distribution $P(y_t|\mathbf{x}_t)$. Instead, we use the class conditional distribution $P(\mathbf{x}_t|y_t)$ to approximate $P(y_t|\mathbf{x}_t)$. Because $P(\mathbf{x}_t|y_t)$ and $P(y_t|\mathbf{x}_t)$ can be quite involved according to the sufficient statistics when sample sizes are large~\cite{long2013transfer}. In order to compute $P(\mathbf{x}_t|y_t)$, we apply prediction on $\mathcal{D}_t$ using some base classifier trained on $\mathcal{D}_s$ to get the soft labels for $\mathcal{D}_t$. The soft labels may be less reliable, so we iteratively refine the them.

In order to compute the marginal and conditional distribution divergences in Eq.~(\ref{equ-mummd}), we adopt maximum mean discrepancy (MMD)~\cite{pan2011domain} to empirically estimate both distribution discrepancies. As a nonparametric measurement, MMD has been widely applied to many existing transfer learning approaches~\cite{pan2011domain,long2013transfer}. Formally speaking, Eq.~(\ref{equ-mummd}) can be represented as
\begin{equation}
\label{equ-mmd-bda}
\Scale[.9]{
	\begin{split}
	D(\mathcal{D}_s,\mathcal{D}_t)
	\approx&(1-\mu)\left \Vert \frac{1}{n} \sum_{i=1}^{n} \mathbf{x}_{s_i} - \frac{1}{m} \sum_{j=1}^{m} \mathbf{x}_{t_j} \right \Vert ^2_\mathcal{H}\\
	&+\mu \sum_{c=1}^{C}\left \Vert \frac{1}{n_c} \sum_{\mathbf{x}_{s_i} \in \mathcal{D}^{(c)}_s} \mathbf{x}_{s_i} - \frac{1}{m_c} \sum_{\mathbf{x}_{t_j} \in \mathcal{D}^{(c)}_t} \mathbf{x}_{t_j} \right \Vert ^2_\mathcal{H}
	\end{split}
}
\end{equation}
where $\mathcal{H}$ denotes the reproducing kernel Hilbert space (RKHS), $c \in \{1,2,\cdots,C\}$ is the distinct class label, $n,m$ denote the number of samples in the source / target domain, and $\mathcal{D}^{(c)}_s$ and $\mathcal{D}^{(c)}_t$ denote the samples belonging to class $c$ in source and target domain, respectively. $n_c=|\mathcal{D}^{(c)}_s|, m_c=|\mathcal{D}^{(c)}_t|$, denoting the number of samples belonging to $\mathcal{D}^{(c)}_s$ and $\mathcal{D}^{(c)}_t$, respectively. The first term denotes the marginal distribution distance between domains, while the second term is the conditional distribution distance. 

By further taking advantage of matrix tricks and regularization, Eq.~(\ref{equ-mummd}) can be formalized as:
\begin{equation}
\label{eqn-bda}
	\begin{split}
	\min ~ &\mathrm{tr}\left(\mathbf{A}^\top \mathbf{X} \left((1-\mu)\mathbf{M}_0 + \mu \sum_{c=1}^{C}\mathbf{M}_c \right) \mathbf{X}^\top \mathbf{A}\right) + \lambda \Vert \mathbf{A}\Vert^2_{F}\\
	\text{s.t.} ~ &
	\mathbf{A}^\top \mathbf{X} \mathbf{H} \mathbf{X}^\top \mathbf{A} = \mathbf{I},\quad 0 \le \mu \le 1
	\end{split}
\end{equation}

There are two terms in Eq.~(\ref{eqn-bda}): the adaptation of marginal and conditional distribution with balance factor (term 1), and the regularization term (term 2). $\lambda$ is the regularization parameter with $\left \Vert \cdot \right \Vert^2_F$ the Frobenius norm. Two constraints are involved in Eq.~(\ref{eqn-bda}): the first constraint ensures that the transformed data ($\mathbf{A}^\top \mathbf{X}$) should preserve the inner properties of the original data. The second constraint denotes the range of the balance factor $\mu$.

More specifically, in Eq.~(\ref{eqn-bda}), $\mathbf{X}$ denotes the input data matrix composed of $\mathbf{x}_s$ and $\mathbf{x}_t$, $\mathbf{A}$ denotes the transformation matrix, $\mathbf{I} \in \mathbb{R}^{(n+m)\times(n+m)}$ is the identity matrix, and $\mathbf{H}$ is the centering matrix i.e. $\mathbf{H}=\mathbf{I}-(1/n)\mathbf{1}$. Similar as in work~\cite{long2013transfer}, $\mathbf{M}_0$ and $\mathbf{M}_c$ are MMD matrices and can be constructed in the following ways:
\begin{equation}
\label{equ-mo}
(\mathbf{M}_0)_{ij}=\begin{cases}
\frac{1}{n^2},  & \mathbf{x}_i,\mathbf{x}_j \in \mathcal{D}_s\\ 
\frac{1}{m^2}, & \mathbf{x}_i,\mathbf{x}_j \in \mathcal{D}_t\\ 
-\frac{1}{mn}, & \text{otherwise} 
\end{cases}
\end{equation}
\begin{equation}
	\label{equ-mc}
	(\mathbf{M}_c)_{ij}=\begin{cases}
	\frac{1}{n^2_c},  & \mathbf{x}_i,\mathbf{x}_j \in \mathcal{D}^{(c)}_s\\ 
	\frac{1}{m^2_c}, & \mathbf{x}_i,\mathbf{x}_j \in \mathcal{D}^{(c)}_t\\ 
	-\frac{1}{m_c n_c}, & \begin{cases}
	\mathbf{x}_i \in \mathcal{D}^{(c)}_s ,\mathbf{x}_j \in \mathcal{D}^{(c)}_t \\ 
	\mathbf{x}_i \in \mathcal{D}^{(c)}_t ,\mathbf{x}_j \in \mathcal{D}^{(c)}_s
	\end{cases}\\
	0, & \text{otherwise}
	\end{cases}
\end{equation}


\textbf{Learning algorithm}: Denote $\Phi=(\phi_1,\phi_2,\cdots,\phi_d)$ as Lagrange multipliers, then Lagrange function for Eq.~(\ref{eqn-bda}) is
\begin{equation}
\label{eqn-lag}
	\begin{split}
	L = ~&\mathrm{tr}\left(\mathbf{A}^\top \mathbf{X} \left((1-\mu)\mathbf{M}_0 + \mu \sum_{c=1}^{C}\mathbf{M}_c \right) \mathbf{X}^\top \mathbf{A}\right) \\
	&+ \lambda \Vert \mathbf{A}\Vert^2_{F} + \mathrm{tr} \left((\mathbf{I}-\mathbf{A}^\top \mathbf{X} \mathbf{H} \mathbf{X}^\top \mathbf{A})\Phi\right)
	\end{split}
\end{equation}
Set derivative $\partial L/\partial \mathbf{A}=0$, the optimization can be derived as a generalized eigendecomposition problem
\begin{equation}
\label{equ-eigen}
	\begin{split}
	\left(\mathbf{X} \left((1-\mu)\mathbf{M}_0 + \mu \sum_{c=1}^{C} \mathbf{M}_c \right)\mathbf{X}^\top + \lambda \mathbf{I} \right) \mathbf{A}
	=\mathbf{X} \mathbf{H} \mathbf{X}^\top \mathbf{A} \Phi
	\end{split}
\end{equation}
Finally, the optimal transformation matrix $\mathbf{A}$ can be obtained by solving Eq.~(\ref{equ-eigen}) and finding its $d$ smallest eigenvectors. 

\textit{Estimation of $\mu$}: Note that $\mu$ is technically not a free parameter like $\lambda$ and it has to be estimated according to data distributions. However, there is no effective solution for its estimation. For now, we evaluate the performance of $\mu$ by searching its values in experiments. For real application, we recommend getting the optimal $\mu$ through cross-validation.

\subsection{Weighted Balanced Distribution Adaptation}
\label{sec-method-wbda}



BDA is able to adaptively leverage the importance of marginal and conditional distributions between domains. BDA indicates when the marginal distributions are relatively close, the performance of transfer learning is highly dependent on the conditional distribution distance. When computing the conditional distributions, BDA utilizes class conditional distributions instead, i.e. $P(\mathbf{x}|y)$ is used to approximate $P(y|\mathbf{x})$. This implicitly assumes that the probability of this class in each domain is similar, which is usually not the case in real world. In this section, we propose a more robust approximation of the conditional distribution for class imbalance problem:
\begin{equation}
\label{equ-wbda-prior}
\begin{split}
&~~~\left\Vert P(y_s|\mathbf{x}_s)-P(y_t|\mathbf{x}_t) \right\Vert ^2_\mathcal{H}\\
&=\left\Vert\frac{P(y_s)}{P(\mathbf{x}_s)} P(\mathbf{x}_s|y_s)-\frac{P(y_t)}{P(\mathbf{x}_t)} P(\mathbf{x}_t|y_t)\right\Vert ^2_\mathcal{H}\\
&=\left\Vert\alpha_s P(\mathbf{x}_s|y_s) - \alpha_t P(\mathbf{x}_t|y_t)\right\Vert ^2_\mathcal{H}
\end{split}
\end{equation}

Technically, we approximate $\alpha_s$ and $\alpha_t$ by the class prior of both domains. To this end, weighted balanced distribution adaptation (W-BDA) is proposed to balance the class proportion of each domain. Evaluating the conditional distribution divergence in Eq.~(\ref{equ-wbda-prior}) requires to estimate the marginal distributions $P(\mathbf{x}_s)$ and $P(\mathbf{x}_t)$. However, it is non-trivial. Since BDA is fully capable of adapting $P(\mathbf{x}_s)$ and $P(\mathbf{x}_t)$, we do not estimate them in this step and assume they are unchanged. Then, we construct a weight matrix $\mathbf{W}_c$ for each class:
\begin{equation}
\label{equ-wc}
	(\mathbf{W}_c)_{ij}=\begin{cases}
	\frac{P\left(y^{(c)}_s\right)}{n^2_c},  & \mathbf{x}_i,\mathbf{x}_j \in \mathcal{D}^{(c)}_s\\ 
	\frac{P\left(y^{(c)}_t\right)}{m^2_c}, & \mathbf{x}_i,\mathbf{x}_j \in \mathcal{D}^{(c)}_t\\ 
	-\frac{\sqrt{P\left(y^{(c)}_s\right)P\left(y^{(c)}_t\right)}}{m_c n_c}, & \begin{cases}
	\mathbf{x}_i \in \mathcal{D}^{(c)}_s ,\mathbf{x}_j \in \mathcal{D}^{(c)}_t \\ 
	\mathbf{x}_i \in \mathcal{D}^{(c)}_t ,\mathbf{x}_j \in \mathcal{D}^{(c)}_s
	\end{cases}\\
	0, & \text{otherwise}
	\end{cases}
\end{equation}
where $P\left(y^{(c)}_s\right)$ and $P\left(y^{(c)}_t\right)$ denote the class prior on class $c$ in the source and target domain, respectively.

Embedding Eq.~(\ref{equ-wc}) into BDA, we get the trace optimization problem of W-BDA:
\begin{equation}
\label{equ-wbda}
\begin{split}
\min ~ &\mathrm{tr}\left(\mathbf{A}^\top \mathbf{X} \left((1-\mu)\mathbf{M}_0 + \mu \sum_{c=1}^{C}\mathbf{W}_c \right) \mathbf{X}^\top \mathbf{A}\right) + \lambda \Vert \mathbf{A}\Vert^2_{F}\\
\text{s.t.} ~ &
\mathbf{A}^\top \mathbf{X} \mathbf{H} \mathbf{X}^\top \mathbf{A} = \mathbf{I},\quad 0 \le \mu \le 1
\end{split}
\end{equation}



\textit{Remark:}
Eq.~(\ref{equ-mo}) of BDA and Eq.~(\ref{equ-wc}) of W-BDA are much similar in spirit. Their differences are: 1) Eq.~(\ref{equ-mo}) of BDA only considers the number of samples in each class, while Eq.~(\ref{equ-wc}) also considers the class prior. 2) Eq.~(\ref{equ-wc}) provides more accurate approximation to the conditional distributions than Eq.~(\ref{equ-mo}) when handling the class imbalance.



\textbf{Kernelization}: When applied to nonlinear problems, we can use a kernel map $\psi$: $\mathbf{x} \mapsto \psi(\mathbf{x})$, and a kernel matrix $\mathbf{K}=\psi(\mathbf{X})^\top \psi(\mathbf{X})$. The kernel matrix $\mathbf{K} \in \mathbb{R}^{(n+m) \times (n+m)}$ can be constructed using linear or RBF kernel.


In summary, Algorithm~\ref{algo-bda} presents the detail of BDA and W-BDA methods.

\begin{algorithm}[!ht] 
	\caption{BDA: Balanced Distribution Adaptation}  
	\label{algo-bda}  
	\renewcommand{\algorithmicrequire}{\textbf{Input:}} 
	\renewcommand{\algorithmicensure}{\textbf{Output:}}
	\begin{algorithmic}[1]  
		\REQUIRE ~~
		Source and target feature matrix $\mathbf{X}_s$ and $\mathbf{X}_t$, source label vector $\mathbf{y}_s$, \#dimension $d$, balance factor $\mu$, regularization parameter $\lambda$\\
		\ENSURE ~~
		Transformation matrix $\mathbf{A}$ and classifier $f$\\
		\STATE Train a base classifier on $\mathbf{X}_s$ and apply prediction on $\mathbf{X}_t$ to get its soft labels $\hat{\mathbf{y}}_t$. Construct $\mathbf{X} = [\mathbf{X}_s,\mathbf{X_t}]$, initialize $\mathbf{M}_0$ and $\mathbf{M}_c$ by Eq.~(\ref{equ-mo}) and~(\ref{equ-mc}) (or $\mathbf{W}_c$ using Eq.~(\ref{equ-wc}) for W-BDA)
		\REPEAT 
		\STATE Solve the eigendecomposition problem in Eq.~(\ref{equ-eigen}) (or Eq.~(\ref{equ-wbda}) for W-BDA) and use $d$ smallest eigenvectors to build $\mathbf{A}$
		\STATE Train a classifier $f$ on $\{\mathbf{A}^\top \mathbf{X}_s,\mathbf{y}_s\}$
		\STATE Update the soft labels of $\mathcal{D}_t$: $\hat{\mathbf{y}}_t=f(\mathbf{A}^\top \mathbf{X}_t)$
		\STATE Update matrix $\mathbf{M}_c$ using Eq.~(\ref{equ-mc}) (or update $\mathbf{W}_c$ using Eq.~(\ref{equ-wc}) for W-BDA)
		\UNTIL{Convergence}  
		\RETURN Classifier $f$  
	\end{algorithmic}  
\end{algorithm}

\section{Experiments}
\label{sec-exp}
In this section, we evaluate the performance of the proposed methods through extensive experiments.


\subsection{Datasets}
\label{sec-exp-data}

We adopt five widely-used datasets: USPS + MNIST, COIL20 and Office + Caltech. Table~\ref{tb-dataset} shows the details of the datasets. \textbf{USPS} (U) and \textbf{MNIST} (M) are standard digit recognition datasets containing handwritten digits from 0-9. USPS consists of 7,291 training images and 2,007 test images. MNIST contains 60,000 training images and 10,000 test images. \textbf{COIL20} (CO) includes 1,440 images belonging to 20 objects. \textbf{Office} is composed of three real-world object domains: \textbf{Amazon}, \textbf{Webcam} and \textbf{DSLR}. It has 4,652 images with 31 object categories. \textbf{Caltech-256} (C) contains 30,607 images and 256 categories. Detailed descriptions about those datasets can be found in~\cite{long2013transfer}. For all the datasets, we follow~\cite{long2013transfer} to construct 16 different tasks.

\begin{table}[h!]
	\centering
	\setlength{\abovecaptionskip}{-.1cm}
	\setlength{\belowcaptionskip}{-1cm}
	\caption{Introduction of the five digit/object datasets.}
	\label{tb-dataset}
	\resizebox{0.48\textwidth}{!}{
		\begin{tabular}{|c|c|c|c|c|c|}
			\hline
			\textbf{Dataset} & \textbf{Type} & \textbf{\#Sample} & \textbf{\#Feature} & \textbf{\#Class} & \textbf{Domain} \\ \hline
			USPS & Digit & 1,800 & 256 & 10 & U \\ \hline
			MNIST & Digit & 2,000 & 256 & 10 & M \\ \hline
			COIL20 & Object & 1,440 & 1,024 & 20 & CO1, CO2 \\ \hline
			Office & Object & 1,410 & 800 & 10 & A, W, D \\ \hline
			Caltech & Object & 1,123 & 800 & 10 & C \\ \hline
		\end{tabular}
	}
\end{table}

\subsection{Comparison Methods}
\label{sec-exp-compmethod}
We choose six state-of-the-art comparison methods:

\begin{itemize}
	\item 1 Nearest Neighbor classifier (1NN)
	\item Principal Component Analysis (PCA) + 1NN
	\item Geodesic Flow Kernel (GFK)~\cite{gong2012geodesic} + 1NN
	\item Transfer Component Analysis (TCA)~\cite{pan2011domain} + 1NN
	\item Joint Distribution Adaptation (JDA)~\cite{long2013transfer} + 1NN
	\item Transfer Subspace Learning (TSL)~\cite{si2010bregman} + 1NN
\end{itemize}

Among those methods, 1NN and PCA are traditional learning methods, while GFK, TCA, JDA, and TSL are state-of-the-art transfer learning approaches. 

\subsection{Implementation Details}
\label{sec-exp-imp}
PCA, TCA, JDA, TSL, and BDA are acting as dimensionality reduction process, then 1NN is applied. For GFK, 1NN is applied after we get the geodesic flow kernel. For BDA and W-BDA, $\mu$ is searched in $\{0,0.1,\cdots,0.9,1.0\}$. Since BDA can achieve a stable performance under a wide range of parameter values, for the comparison study, we set $d=100$; $\lambda=0.1$ for MNIST + USPS / Office + Caltech datasets and $\lambda=0.01$ for COIL20 dataset. For the kernel-based methods, we use linear kernel. The iteration number for JDA and TCA is set to be $T=10$. The codes of BDA and W-BDA are available online\footnote{Code available at \url{http://tinyurl.com/yd3ol4om}}. Classification \textit{accuracy} on target domain is adopted as the evaluation metric which is widely used in literatures~\cite{long2013transfer,yan2017mind}.

\begin{table}[h!]
	\centering
	\setlength{\abovecaptionskip}{-.1cm}
	\setlength{\belowcaptionskip}{-1cm}
	\caption{Accuracy (\%) of BDA and other methods on 16 tasks.}
	\label{tb-result}
	\resizebox{0.49\textwidth}{!}{
		\begin{tabular}{c|ccccccc}
			\hline
			Task & 1NN & PCA & GFK & TCA & JDA & TSL & BDA \\ \hline \hline
			U $\rightarrow$ M & 44.70 & 44.95 & 46.45 & 52.20 & 57.45 & 53.75 & \textbf{59.35} \\ \hline
			M $\rightarrow$ U & 65.94 & 66.22 & 67.22 & 54.28 & 62.89 & 66.06 & \textbf{69.78} \\ \hline
			CO1 $\rightarrow$ CO2 & 83.61 & 84.72 & 72.50 & 88.61 & 97.22 & 88.06 & \textbf{97.22} \\ \hline
			CO2 $\rightarrow$ CO1 & 82.78 & 84.03 & 74.17 & 96.25 & 86.39 & 87.92 & \textbf{96.81} \\ \hline
			C $\rightarrow$ A & 23.70 & 36.95 & 41.02 & 44.89 & 42.90 & 44.47 & \textbf{44.89} \\ \hline
			C $\rightarrow$ W & 25.76 & 32.54 & 40.68 & 36.61 & 38.64 & 34.24 & \textbf{38.64} \\ \hline
			C $\rightarrow$ D & 25.48 & 38.22 & 38.85 & 45.86 & 47.13 & 43.31 & \textbf{47.77} \\ \hline
			A $\rightarrow$ C & 26.00 & 34.73 & 40.25 & 40.78 & 38.82 & 37.58 & \textbf{40.78} \\ \hline
			A $\rightarrow$ W & 29.83 & 35.59 & 38.98 & 37.63 & 37.29 & 33.90 & \textbf{39.32} \\ \hline
			A $\rightarrow$ D & 25.48 & 27.39 & 36.31 & 31.85 & 40.13 & 26.11 & \textbf{43.31} \\ \hline
			W $\rightarrow$ C & 19.86 & 26.36 & 30.72 & 27.16 & 25.29 & \textbf{29.83} & 28.94 \\ \hline
			W $\rightarrow$ A & 22.96 & 31.00 & 29.75 & 30.69 & 31.84 & 30.27 & \textbf{32.99} \\ \hline
			W $\rightarrow$ D & 59.24 & 77.07 & 80.89 & 90.45 & 90.45 & 87.26 & \textbf{91.72} \\ \hline
			D $\rightarrow$ C & 26.27 & 29.65 & 30.28 & 32.50 & 30.99 & 28.50 & \textbf{32.50} \\ \hline
			D $\rightarrow$ A & 28.50 & 32.05 & 32.05 & 31.52 & 32.25 & 27.56 & \textbf{33.09} \\ \hline
			D $\rightarrow$ W & 63.39 & 75.93 & 75.59 & 87.12 & 91.19 & 85.42 & \textbf{91.86} \\ \hline \hline
			Average & 40.84 & 47.34 & 48.48 & 51.78 & 53.18 & 50.27 & \textbf{55.56} \\ \hline
		\end{tabular}
	}
\end{table}

\begin{figure}[h!]
	\centering
	\setlength{\abovecaptionskip}{-.3cm}
	\hspace{.1in}
	\subfigure[MMD distance]{
		\includegraphics[scale=0.26]{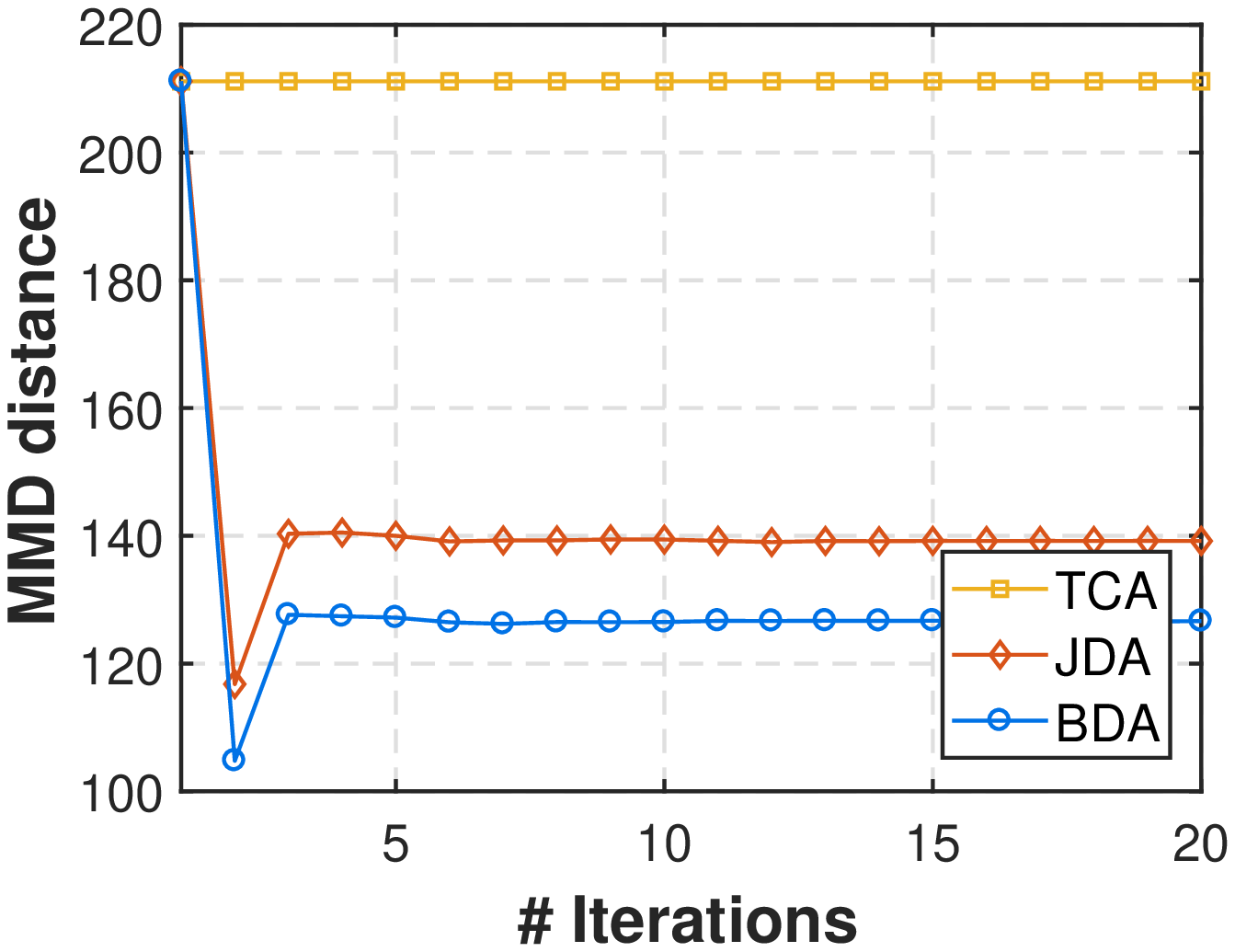}
		\label{fig-sub-mmd}}\hfill
	\subfigure[Accuracy]{
		\includegraphics[scale=0.26]{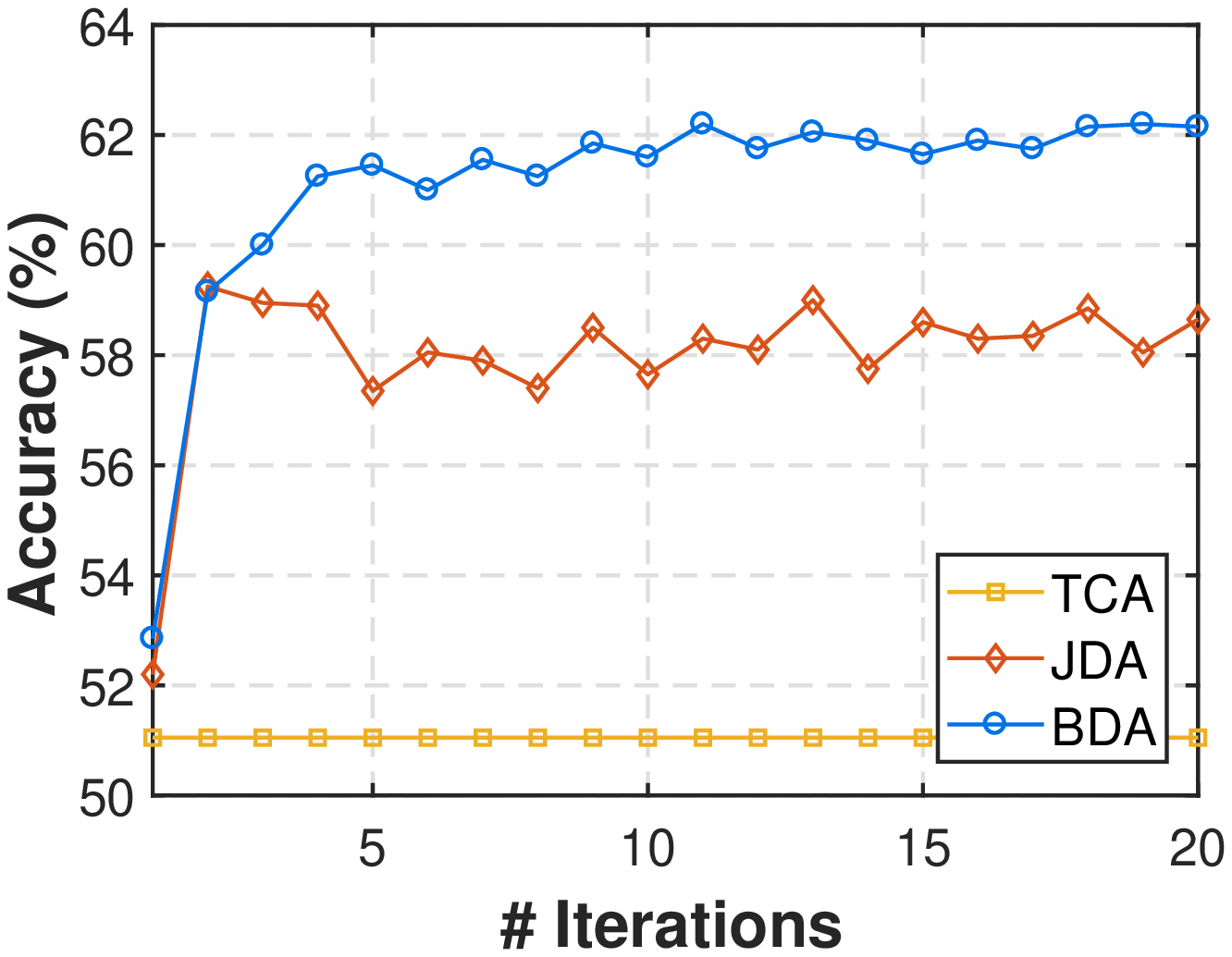}
		\label{fig-sub-acc}}
	\caption{MMD distance and classification accuracy comparison of TCA, JDA and BDA on U $\rightarrow$ M. It can be noted that BDA achieves better accuracy with relatively small MMD distance.}
	\label{fig-effect}
	\vspace{-.5cm}
\end{figure}

\subsection{Performance Evaluation of BDA}
\label{sec-exp-performance}
\subsubsection{Classification accuracy}
We test the performance of BDA and the other comparison methods on 16 cross-domain learning tasks. The results are shown in Table~\ref{tb-result}, based on which, we can draw the following observations.

First, BDA outperforms most of the existing methods~(15 out of 16 tasks). Specifically, the average classification accuracy of BDA is \textbf{55.56\%}, which shows an average improvement of 2.38\% compared to the best comparison method JDA. JDA is only capable of adapting the marginal and conditional distribution with the equal weight ($\mu=0.5$). Thus JDA can be considered as a special case of BDA. However, BDA can dramatically improve the accuracy by adjusting the balance parameter $\mu$ to adapt various scenarios.

Second, TCA is also a special case of BDA ($\mu = 0$) since it only adapts the marginal distribution. Therefore, the performance of TCA is worse than JDA and BDA. 


Third, TSL only adapts the marginal distributions which highly relies on the distribution density. The performance of GFK is better on object recognition tasks. The reason is that GFK learns a global geodesic flow kernel on the low-dimension representation, which may be enough to transit smoothly for the object datasets. But as for the digit tasks, it may not enough to construct smooth transit when the marginal distribution distance is large.

Last, all transfer learning methods perform better than traditional learning approaches due to the large distribution gap between different domains. This indicates the effectiveness of transfer learning methods, among which BDA could achieve the best performance.

\subsubsection{Effectiveness of distribution adaptation} 
We further verify the effectiveness of BDA by comparing its distribution adaptation with other two distribution adaptation methods: TCA and JDA. Specifically, we investigate their performance with MMD distances calculated using Eq.~(\ref{eqn-bda}).

Fig.~\ref{fig-sub-mmd} and Fig.~\ref{fig-sub-acc} show the MMD distance and accuracy of TCA, JDA, and BDA with increasing iteration, respectively. Based on the results, we can observe: a) MMD distances of all methods can be reduced. This indicates the effectiveness of TCA, JDA and BDA; b) MMD distance of TCA is not reduced largely as it only adapts the marginal distribution distance and requires no iteration; c) MMD distance of JDA is obviously larger than BDA, since BDA could balance the importance of marginal and conditional distribution via $\mu$; d) BDA achieves the best performance. 

\begin{figure}[t!]
	\vspace{-.5cm}
	\centering
	\setlength{\abovecaptionskip}{-.1cm}
	\setlength{\belowcaptionskip}{-1cm}
	\includegraphics[scale=0.4]{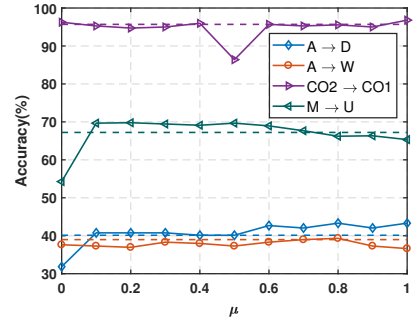}
	\caption{Classification accuracy w.r.t. $\mu$ on different tasks. Dashed lines are the best comparison methods.}
	\label{fig-mu}
	\vspace{-.5cm}
\end{figure}

\subsection{Effectiveness of Balance Factor}
\label{sec-exp-param}
In this section, we evaluate the effectiveness of the balance factor $\mu$. We run BDA with $\mu \in \{0,0.1,\cdots,1.0\}$ on some tasks and compare the performances with the best baseline method. Fig.~\ref{fig-mu} shows the results. It is obvious that the optimal $\mu$ varies on different tasks, indicating the importance to balance the marginal and conditional distributions between domains. In comparison, the best baseline JDA (dash lines) is only the special case of BDA ($\mu = 0.5$), which means to treat those distributions equally. However, this assumption does not hold. In task A $\rightarrow$ W with optimal $\mu = 0.8$, it means the marginal distributions are almost the same so the performance of transfer learning mostly depends on conditional distributions. In task M $\rightarrow$ U with optimal $\mu = 0.1$, it means the marginal distributions contribute most to the discrepancy, so $\mu$ is relatively small. In other 13 tasks, the observations are similar. It indicates in cross-domain learning problems, $\mu$ is extremely important to balance both the marginal and conditional distributions. Therefore, BDA is more capable of achieving good performance.

To be noticed, there may be more than one optimal $\mu$ for some tasks (A $\rightarrow$ D), and the tendency of $\mu$ is not always stable (CO2 $\rightarrow$ CO1). The problems behind those facts still need to be addressed in future research.

\subsection{Effectiveness of Weighted BDA}
\label{sec-exp-wbda}
We extensively verify the effectiveness of proposed W-BDA. We choose some tasks with highly imbalanced class distributions and compare the performance of W-BDA with BDA and JDA. TABLE.~\ref{tb-wbda} demonstrates the classification accuracy of 6 tasks. Note that for comparison,  classes on tasks 1 $\sim$ 4 are rather imbalanced, while classes are rather balanced on tasks 5 $\sim$ 6.

\vspace{-.3cm}
\begin{table}[h!]
	\setlength{\abovecaptionskip}{-.1cm}
	\setlength{\belowcaptionskip}{-.5cm}
	\centering
	\caption{Accuracy of JDA, BDA and W-BDA on some tasks.}
	\label{tb-wbda}
	\begin{tabular}{|c|c|c|c|c|}
		\hline
		Index & Task & JDA & BDA & WBDA \\ \hline
		1 & C $\rightarrow$ D & 47.13 & 47.77 & \textbf{48.41} \\ \hline
		2 & W $\rightarrow$ C & 25.29 & 28.94 & \textbf{31.08} \\ \hline
		3 & U $\rightarrow$ M & 57.45 & 59.15 & \textbf{59.35} \\ \hline
		4 & A $\rightarrow$ W & 37.29 & 39.32 & \textbf{40.68} \\ \hline
		5 & C $\rightarrow$ A & 42.90 & 44.89 & \textbf{45.20} \\ \hline
		6 & CO1 $\rightarrow$ CO2 & 97.22 & \textbf{97.22} & 96.69 \\ \hline
	\end{tabular}
\end{table}
\vspace{-.3cm}
From the results, we can observe: 1) JDA achieves the worst results since it does not consider the gap between marginal and conditional distributions. BDA is able to handle the distribution discrepancy and outperforms JDA in all situations. 2) For the first four tasks which are under imbalanced class distributions, W-BDA could improve the performance by adaptively weighting each class. In the other two tasks where class distributions are rather balanced, W-BDA still achieves comparable results. On the other 10 tasks, the results follow the same tendency. To sum up, the results demonstrate that in transfer learning, W-BDA remains an effective method to balance the different class distribution between domains. 

In addition, BDA and W-BDA have other two parameters: feature dimension $d$ and regularization parameter $\lambda$. Their sensitivity evaluations are omitted due to page limit. In our actual experiments, BDA and W-BDA are relatively robust to those two parameters.




\section{Conclusion}
\label{sec-con}
Balancing the probability distributions and class distributions between domains are both two important problems in transfer learning. In this paper, we propose Balanced Distribution Adaptation (BDA) to adaptively weight the importance of both marginal and conditional distribution adaptations. Thus, it could significantly improve the transfer learning performance. Moreover, we consider handling the class imbalance problem for transfer learning by proposing Weighted BDA (W-BDA). Extensive experiments on five image datasets demonstrate the superiority of our methods over several state-of-the-art methods. In the future, we will continue the exploration in these two aspects: by developing more strategies to leverage the distributions and handle the class imbalance problem.

\vspace{-.8cm}
\section*{Acknowledgment}
This work is supported in part by National Key R \& D Plan of China (No.2016YFB1001200), NSFC (No.61572471), and Beijing Municipal Science \& Technology Commission (No.Z161100000216140 \& Z171100000117013).



\vspace{-.18cm}
\newcommand{\BIBdecl}{\setlength{\itemsep}{0.1 em}}
\bibliographystyle{IEEEtran}
\bibliography{icdm17}
%

\end{document}